\documentclass[sigconf]{acmart}
\usepackage[normalem]{ulem}
\usepackage{svg}
\usepackage{subfigure}
\usepackage{multirow}
\usepackage{amsmath}

\AtBeginDocument{%
  \providecommand\BibTeX{{%
    \normalfont B\kern-0.5em{\scshape i\kern-0.25em b}\kern-0.8em\TeX}}}

\setcopyright{acmlicensed}
\copyrightyear{2018}
\acmYear{2018}
\acmDOI{XXXXXXX.XXXXXXX}

\acmConference[Conference acronym 'XX]{Make sure to enter the correct
  conference title from your rights confirmation emai}{June 03--05,
  2018}{Woodstock, NY}
\begin{document}

\title{Knowledge Probing for Graph Representation Learning}



\author{Mingyu Zhao}
\authornote{Both authors contributed equally to this research.}
\email{my.zhao1@siat.ac.cn}
\affiliation{%
  \institution{School of Cyber Science and Technology, Shenzhen Campus of Sun Yat-sen University. Shenzhen Institute of Advanced Technology, Chinese Academy of Sciences, University of Chinese Academy of Sciences.}
  \city{Shenzhen}
  \state{Guangdong}
  \country{China}
}

\author{Xingyu Huang}
\authornotemark[1]
\email{xingyuhuang1998@outlook.com}
\affiliation{%
  \institution{School of Cyber Science and Technology, Shenzhen Campus of Sun Yat-sen University. Shenzhen Institute of Advanced Technology, Chinese Academy of Sciences.}
  \city{Shenzhen}
  \state{Guangdong}
  \country{China}
}

\author{Ziyu Lyu}
\authornote{Corresponding author.}
\email{luziyucrystal@163.com}
\affiliation{%
  \institution{School of Cyber Science and Technology, Sun Yat-sen University}
  \city{Shenzhen}
  \state{Guangdong}
  \country{China}
}

\author{Yanlin Wang}
\email{yanlin-wang@outlook.com}
\affiliation{%
  \institution{Sun Yat-sen University}
  \city{Guangzhou}
  \state{Guangdong}
  \country{China}
}

\author{Lixin Cui}
\email{cuilixin@cufe.edu.cn}
\affiliation{%
  \institution{Central University of Finance and Economics}
  \city{Beijing}
  \country{China}
}

\author{Lu Bai}
\email{bailu@bnu.edu.cn}
\affiliation{%
  \institution{Beijing Normal University}
  \city{Beijing}
  \country{China}
}

\begin{abstract}
Graph learning methods have been extensively applied in diverse application areas. However, what kind of inherent graph properties e.g. graph proximity, graph structural information has been encoded into graph representation learning for downstream tasks is still under-explored. In this paper, we propose a novel graph probing framework (GraphProbe) to investigate and interpret whether the family of graph learning methods has encoded different levels of knowledge in graph representation learning. Based on the intrinsic properties of graphs, we design three probes to systematically investigate the graph representation learning process from different perspectives, respectively the node-wise level, the path-wise level, and the structural level. We construct a thorough evaluation benchmark with nine representative graph learning methods from random walk based approaches, basic graph neural networks and self-supervised graph methods, and probe them on six benchmark datasets for node classification, link prediction and graph classification. The experimental evaluation verify that GraphProbe can estimate the capability of graph representation learning. Remaking results have been concluded: GCN and WeightedGCN methods are relatively versatile methods achieving better results with respect to different tasks.
\end{abstract}

\ccsdesc[500]{Information systems}
\ccsdesc[500]{Computing methodologies~Knowledge representation and reasoning}




\keywords{Graph Representation Learning, Knowledge Probing, Evaluation}




\maketitle
\newcommand\info{I}
\newcommand\probe{P}
\newcommand{\ziyu}[1]{{\textcolor{red}{[ziyu: #1]}}}
\newcommand{\mingyu}[1]{{\textcolor{blue}{[mingyu: #1]}}}
\newcommand\graph{G}
\newcommand\nodeset{V}
\newcommand\node{v}
\newcommand\edgeset{E}
\newcommand\edge{e}
\newcommand\adjmat{A}
\newcommand\adj{a}
\newcommand\nodefeat{\mathbf{X}}
\newcommand\nodeemb{\mathbf{h}}
\newcommand\graphemb{\mathbf{H}}
\newcommand\simmatrix{\mathbf{S}}

\newcommand\model{M}
\newcommand\task{T}
\newcommand\dataset{D}

\newcommand{\param}{W}
\newcommand\wlabel{WL}

\newcommand\category{L}
\newcommand\centroid{k}
\newcommand\cluster{C}
\newcommand\sample{e}

\newcommand\batch{b}

\def\halfcheckmark{\ding{51}\textsuperscript{\kern-0.5em\normalsize\ding{55}}}

\section{Introduction} 
\label{intro}
Graphs are a prevalent data structure and have been broadly in multiple fields \cite{Hamilton2017Representation}. For example, social networks\cite{mccallum2000cora, giles1998citeseer}, molecular graph structures, and biological protein networks are universally modeled as graphs \cite{Borgwardt05:proteins}. In recent decades, a lot of graph representations learning methods have been devised, ranging from matrix factorization methods\cite{Ahmed2013MF,Cao2015GraRep}, random-walk based algorithms\cite{perozzi14:deepwalk, grover16:node2vec}, to the popular family of graph neural networks (GNN) \cite{defferrard16:cheby, kipf17:gcn, velickovic17:gat,  hamilton17:sage, xu18:gin, He20:lightgcn, zhu21:ssgcn}. The graph representation learning methods have demonstrated different performance on the classical downstream tasks, e.g. node classification, link prediction and graph classification. And the diverse graph representation learning methods have been extensively applied in multiple application areas, e.g. social network analysis \cite{mccallum2000cora, giles1998citeseer}, recommender system \cite{sen2008ngcf, Harper15:movie, He20:lightgcn}, and protein classification \cite{Debnath91:mutag, Borgwardt05:proteins}. 

Those graph representation learning methods tend to learn a mapping which embed nodes or (sub)graphs into a low-dimensional vectors by encoding relational information and structural information, and the learned embeddings are used for further downstream tasks \cite{pimentel2020information}. However, there is no study to investigate and explain what kinds of graph properties have been actually coded in the learned embedding through different graph representation learning methods. It lacks a systematical evaluation to probe whether the graph inherent properties (e.g. graph proximity, graph structural information) are encoded into the learned node and graph representations with the popular but black-box graph representation learning methods.

In this paper, we devise a systematic graph probing benchmark (GraphProbe) to investigate what types of knowledge are encoded into graph representation learning for 9 representative graph learning methods from diverse categories including random walk based methods, classical graph neural networks like graph convolution networks (GCN) and graph attention networks (GAT), unsupervised graph learning methods and weighted graph learning methods. We devise three types of probes at three different levels, respectively the \textbf{node-wise} , \textbf{path-wise} and \textbf{structure-wise} levels. \textbf{First}, the node-wise probe is proposed to investigate whether the node-wise influences are encoded in graph representation learning, and two intrinsic node centrality metrics (e.g. eigenvector centrality and betweenness centrality) are adopted to measure node-wise influences. \textbf{Second}, a distance probe is designed to explore whether the path-wise information (distance between two nodes) can be embedded into representation learning of nodes, based on the shortest paths of a pair of nodes. \textbf{Third}, we leverage Weisfeiler-Lehman kernel algorithm \cite{Shervashidze2011WLkernel} as structural evaluation metrics and devise a parameter-free structural probe to interpret whether the structural information is enoverviecoded into the learned graph representations. We conduct extensive experiments to probe the performance of graph representation learning on 6 benchmark datasets from diverse domains for three traditional downstream tasks. Our systematic evaluation has concluded some remarked results.
The main contributions of this paper are summarized as follows:
\begin{itemize}
    \item To our best knowledge, our proposed method is the first time to explore knowledge probing on various types of graph learning models across all classical downstream tasks. A systematic graph probing framework is novelly introduced, in which three types of probes based on graph intrinsic properties are devised respectively from the node-wise, path-wise and structure-wise levels.
 \item  An evaluation benchmark for graph representation learning on node classification, link prediction and graph classification, broadly including nine representative graph learning models from four different categories and six benchmarks datasets from three domains including citation networks, social networks and Bio-chemical networks.
\item Our experimental results conclude remarked findings. Especially, our devised knowledge probes are verified to reflect the capability of graph representation learning, and have competitive and consistent results with the traditional evaluation metrics such as accuracy, AUC and F1 scores. 
\end{itemize}

\section{Related Work} 
\label{related}
\subsection{Graph Representation Learning}
Graph representation Learning methods map structural graph data into low-dimensional dense vectors, by capturing the graph topology structure, node-to-node relationships and other relevant information.  
Early methods for learning representations for nodes on graph-structured data were mainly matrix factorization methods based on dimension reduction \cite{Ahmed2013MF, Cao2015GraRep}, and random walk approaches based on random walk statistics (e.g. DeepWalk \cite{perozzi14:deepwalk} and Node2Vec \cite{grover16:node2vec}). For example, DeepWalk\cite{perozzi14:deepwalk} was the first to input random walk paths into a skip-gram model to learn graph node embeddings. Node2vec\cite{grover16:node2vec} combined both breadth-first and depth-first walks to approximate with negative sampling. However, matrix factorization and random walk methods are shallow embedding approaches \cite{Hamilton2017Representation} and they cannot capture structural similarity. 
In recent decades, graph neural networks have been extensively proposed for graph representation learning \cite{Hamilton2017Representation}. The family of graph neural networks rely on the neighborhood aggregation strategy, and solve the main limitations of the direct encoding methods like matrix factorization and random walk approaches. For example, Kipf et al. \cite{kipf17:gcn} proposed a scalable approach (Graph convolution Network, GCN) for semi-supervised learning on graph-structured
data, by utilizing an efficient layer-wise propagation rule that is based on a first-order approximation of spectral convolutions on graphs. GraphSAGE \cite{hamilton17:sage} was a general inductive graph representation learning framework, and learned a function that generates embeddings by sampling and leveraged node feature information to efficiently generate node embeddings for previously unseen data. In addition, some unsupervised graph learning frameworks have been devised. For example, Variational Graph Auto-Encoders (VGAE) proposed an unsupervised learning framework based on the variational auto-encoder. Based on the recent contrastive learning techniques, You et al. \cite{you:2020:GCL} proposed a graph contrastive learning framework for unsupervised representation learning of graph data and explored four types of graph augmentations to incorporate various priors. Although extensive graph representation learning methods have been proposed and worked on the traditional downstream tasks, no studies have investigated whether the inherent graph structural and topological information has been encoded into graph representation learning. In this paper, we propose a graph knowledge probing framework to probe the latent knowledge within graph representation learning, and answer what kind of graph information have been encoded when performing downstream tasks.

\subsection{Knowledge Probe}
Recently, knowledge probes have been proposed to probe knowledge in pre-trained language models (PLMs) such as ELMo \cite{peters-etal-2018-deep} and BERT \cite{devlin-etal-2019-bert}. Probing methods are designed to understand and interpret what knowledge have been learned in the pre-trained language models, and they probe specific knowledge including linguistic knowledge \citet{shi2016Syntax, conneau2018cram, hewitt2019structural,hou2021birdeyes}, and factual knowledge \citet{petroni2019LAMA}. For example, Hewitt et al. \cite{hewitt2019structural} proposed a structural probe to evaluate whether syntax trees have been encoded in a linear transformation of a neural network's word representation space. The probing results demonstrated that the transforms exist for the two PLMs ELMo and BERT.
Petroni et al. proposed a LAMA benchmark to probe factual knowledge in PLMs using promt-based retrieval.  
The most similar work is \cite{akhondzadeh2023probing} in which a probing framework has been proposed for quantify the chemical knowledge and molecular properties in graph representations for graph based neural networks. Different from previous studies, we propose a holistic graph probing benchmark to understand and interpret whether different types of inherent graph properties have been encoded into graph representation learning, from the node-wise, path-wise and structural-wise levels. The studied graph models covering a broad range of graph learning methods, ranging from random walk based methods to graph neural networks. In addition, we benchmark our knowledge probes on three classical downstream tasks with nine representative graph learning methods. 



\section{The Graph Embedding Probes} \label{method}
\begin{figure*}
    \centering
    \includegraphics[width=0.95\textwidth]{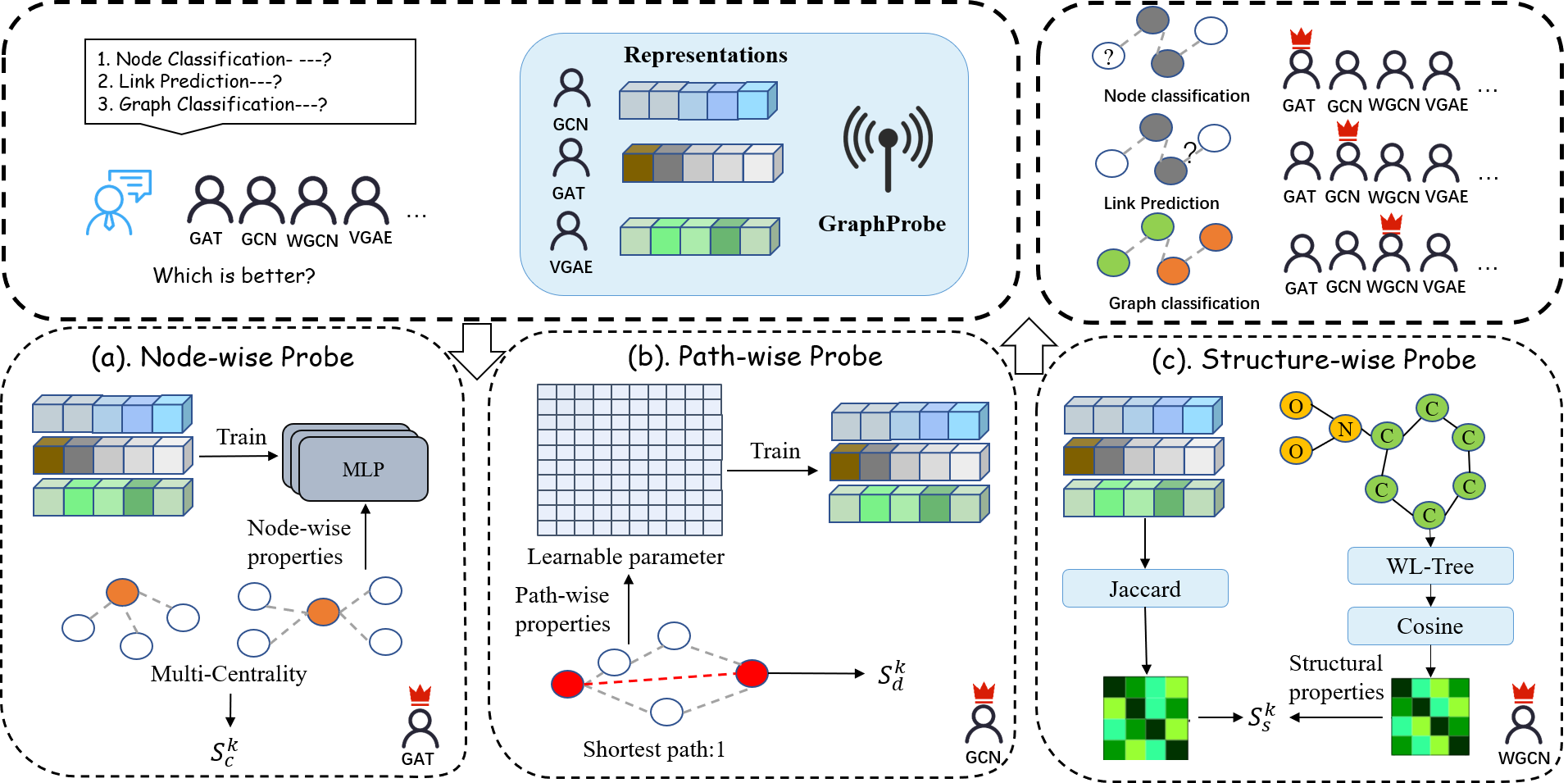}
    \caption{The overall architecture of our proposed GraphProbe.}
    \label{fig:Overview}
    \Description{A diagram showing the architecture of GraphProbe, detailing its various components and their interconnections.}
\end{figure*}
In order to investigate whether the inherent graph properties have been encoded into graph representation learning and reveal why different graph learning methods have different performance on downstream tasks, we we propose a knowledge probing framework (\textbf{GraphProbe}) to probe graph representation learning, and devise three knowledge probes from different levels, respectively \textbf{node-wise}, \textbf{path-wise}, and \textbf{structure-wise} levels. Figure~\ref{fig:Overview} shows the overall architecture of our proposed GraphProbe. From the node-wise level, we devise the node centrality probes to investigate whether the influences of node properties and the local neighbourhood information can be encoded into representation learning of nodes for downstream tasks such as node classification and link prediction. From the path-wise level, we leverage the distance metrics of paths between two nodes to explore whether the path-wise topological information can be encoded into graph representation learning. From the structure-wise level, a graph structural probe is devised to investigate whether the structural information e.g. sub-graph information is encoded into graph representation learning when performing the classical graph classification task. In the following parts, we firstly have a formal problem definition for the knowledge probing on graph representation leaning and then illustrate different probes in details.

\paragraph{\textbf{Problem Definition}} We give a formal definition for the knowledge probing problem on graph representation learning. Given the constructed graph data $\graph=\{\nodeset,\edgeset\}$, $\nodeset$ is the set of nodes in $\graph$, and $\edgeset$ is the set of the edges. $\nodeemb_i$ represents the feature representation of each node $\node_i$, and the dimension of the feature representation is $d$.  The node  representation $\nodefeat$ can be randomly initialized or initialized with meta features. With a graph-based model $\model^k$,  we can obtain the learned node feature representation $\nodeemb^k_{1:|\nodeset|}=\model(\graph,\nodefeat,\theta^k)$. The graph probe $\probe$ is  a function to estimate whether the learned representations encode the specified properties \info,  as defined in Equation~\ref{eqn:probe}.
\begin{equation}
    S^k = \probe (\nodeemb^k_{1:|\nodeset|},\info,\task).
\label{eqn:probe}
\end{equation}
in which $\info$ denotes the investigated metrics used in the devised probes, and $\task$ denotes the applied downstream task.
The probe score $S^k$ estimates how well the learned representations from the graph-based model $\model^k$ encode information for the downstream task $\task$ with respect to $\info$.  In the following sections, we will describe the different probes in details.

\subsection{The Node Centrality Probe}
From the node-wise perspective,  we leverage  the node centrality properties of graph data as the estimated metrics $\info$ because the node centrality reflects the influence and importance of a node and to the extent captures the neighborhood information of a node \cite{das2018study}.  We devise a node centrality probe to investigate whether the learned representations have encoded the node centrality when performing classical downstream tasks, e.g, node classification and link prediction.

We devise a supervised probe to compare the node centrality of two different nodes $\node_i$ and $\node_j$. We calculate the node centrality values $C(\node_i)$ and $C(\node_j)$ based on some graph node centrality metrics (e.g. eigenvector centrality and betweeness centrality), and map the pair-wise centrality comparison into a binary value $l_{ij}$ as in Equation~\ref{eqn:clabel}. The binary value $l$ is used as the centrality label to train the node centrality probe $\probe^c(\nodeemb^k_{1:|\nodeset|},l,\task)$.
\begin{equation}
\label{eqn:clabel}
    l_{ij}=\left\{
\begin{aligned}
1 &   &C(\node_i) \ge C(\node_j) \\
0 &   &C(\node_i) <   C(\node_j). \\
\end{aligned}
\right.
\end{equation}
Following the probe architecture  in \citet{pimentel2020information} which has been designed to reduce the information loss, we adopt a simple learning network two-layer perception (MLP)\footnote{We use a two-layer MLP as the learning network, and the activation function is ReLU.} to learn the supervised probe:
$\probe_c^k(\nodeemb_i^k, \nodeemb_{j}^k) = MLP(\nodeemb_{i}^k \Vert \nodeemb_j^k)$, and output  the probability $p$ that the previous node $\node_i$ has a larger centrality than node $\node_j$. 
Cross-entropy is used as the loss function to train the supervised probe. 
Finally, the probe scores of the node centrality probe $ S^k_c$ for model $\model^k$ are the evaluation measure scores to measure the prediction from the probe based on the learned representations of the model $\model^k$ as follows:
\begin{equation}
    S^k_c =Eval(\probe_c^k(\nodeemb_i^k, \nodeemb_{j}^k),l_{ij}).
\end{equation}
Classical evaluation metrics can be used for $Eval$(e.g. F1-score, AUC, Accuracy). We report results with Accuracy and F1-score \cite{Rijsbergen1979} in experiment section. A higher probe score $S^k_c$ means the graph-based model $\model^k$ has greater ability to encode centrality information into the node representations. We explore two node centrality metrics for $C(\cdot)$, respectively eigenvector centrality \cite{bonacich1987power} and betweeness centrality \cite{Shaw54}.



\paragraph{Eigenvector Centrality}
Eigenvector centrality\cite{bonacich1987power} measures the importance of nodes in a network by exploiting adjacency and eigenvector matrices.
Eigenvector centrality is a unique measure that satisfies certain natural principles for a ranking algorithm\cite{Altman05}. And \citet{Wang22} show that several recommendation algorithms based on node importance have been enhanced with the introduction of eigenvector centrality.
\(\mathbf{A}\in\mathbb{R}^{n\times n}\) is the adjacency matrix such that \(a_{ij} = 1\) if node \(i\) is connected to node \(j\) and \(a_{ij} = 0\) if not. The formal definition of the eigenvalue \(\lambda\) and the eigenvector \(\mathbf{x}\) is \(A\mathbf{x} = \lambda\mathbf{x}\) 
And the principal eigenvector \(\mathbf{x}^p=(x^p_1, \cdots, x^p_n)\) is the eigenvector corresponding to the eigenvalue with the largest modulus. The eigenvector centrality of node \(i\) can be computed as in Equation~\ref{eqn:EC}:
\begin{equation}
\label{eqn:EC}
    EC(i) = x^p_i,
    EC(n_i) = \frac{1}{\lambda}\sum_{n_j\in N(n_i)}EC(n_j),
     x_i = \frac{1}{\lambda}A_{i,j}^Tx_j.
\end{equation}
where $EC(n_i)$, $EC(n_j)$ is the amount of influence that node $n_i$, $n_j$ carries, $N(n_i)$ is the set of direct neighbors of node $n_i$, and $\lambda$ is a constant.

\paragraph{Betweenness Centrality}
In graph theory, betweenness centrality\cite{Freeman1977Centrality} is a measure of centrality in a graph. For every pair of vertices in a connected graph, there exists at least one shortest path between the vertices such that either the number of edges that the path passes through (for unweighted graphs). The betweenness centrality for each vertex is the number of these shortest paths that pass through the vertex. It applies to a wide range of problems in network theory, including problems related to social networks, biology, transport and scientific cooperation\cite{Freeman1977Centrality}.
The betweeness centrality is defined as in Equation~\ref{eqn:BC} :
\begin{equation}
\label{eqn:BC}
    BC(\node_i) = \sum_{\node_i\neq \node_j\neq \node_t \in \nodeset}{\frac{\sigma_{jt}(\node_i)}{\sigma_{jt}}}.
\end{equation}
$\sigma_{jt}$ is number of the shortest paths between node $\node_j$ and $\node_t$, and $\sigma_{jt}(\node_i)$ is the number of shortest path passing through node $\node_i$.

\subsection{The Distance Probe}
In graph data, the distance between two nodes can be estimated by the shortest path. From the path-wise perspective, we devise the distance probe to investigate whether the node representations encode the path-level distance information of graph structure.
Following \citet{hewitt2019structural}, we devise the distance probe $\probe_d$ as to estimate the differences between the grounded shortest paths of two nodes and the vector distance of two nodes' representations. Firstly, we define a family of inner products, $\nodeemb^T\param\nodeemb$ parameterized by any positive semi-definite, the symmetric matrix $\param \in S_{+}^{m\times m}$. Equivalently, we can view this as a linear transformation $B \in \mathbb{R}^{k\times m}$, such that $\param = B^TB$. The inner product $\nodeemb^T\param\nodeemb$ is then equivalent to $(B\nodeemb)^T(B\nodeemb)$, the norm of $\nodeemb$ once transformed by $B$. Every inner product corresponds to a distance metric. Therefore, the definition of the distance between two nodes' embedding $\nodeemb^k_i,\nodeemb^k_j$ is: 

\begin{equation}
    d_B(\nodeemb^k_i,\nodeemb^k_j)^2 = (B(\nodeemb^k_i-\nodeemb^k_j))^T(B(\nodeemb^k_i-\nodeemb^k_j)).
\end{equation}
The distance probe $\probe_d$ is trained to recreate the graph distance of node pairs in the training graph, and optimized through gradient descent in Equation~\ref{eqn:distanceprobe}.
The parameters of our probe are exactly the matrix $B$, which we train to recreate the graph distance of node pairs in the training graph. Specifically, we approximate through gradient descent: 
\begin{equation}
    \mathop{min}\limits_B\sum_{i,j\in \graph}{\Big|d_\graph(n_i, n_j)-d_B(\nodeemb_i, \nodeemb_j)^2\Big|}.
\label{eqn:distanceprobe}
\end{equation}
where $d_G(n_i, n_j)$ denotes the distance of the shortest path of two nodes $\node_i,\node_j$ in $\graph$\footnote{Considering the computational complexity, we only keep the node pairs with distance less than or equal to 3.}.

\begin{equation}
    S^k_d = \frac{1}{\sum_{i,j\in \graph}{\Big|d_\graph(n_i, n_j)-d_B(\nodeemb^k_i, \nodeemb^k_j)^2\Big|}}.
\end{equation}
The probe score $S^k_d$ represents the performance of the probe to recreate the graph distance. Therefore, Bigger $S^k_d$ indicates better performance. 


\subsection{The Graph Structural Probe}
We propose a structural probe to estimate whether the structural information has been encoded into the embedding of the entire graph. The graph representation $\graphemb^k$ is constructed by aggregating the node representations through the readout operation:
\begin{equation}
    \graphemb^k = readout(\graph,\nodeemb^k_{1:|V|}).
\end{equation}
The readout operation can obtain graph-level representation, e.g. sum, mean and max pooling\footnote{The three operations have been implemented in the benchmark}. We report the results with sum operation.

In order to extract the inherent structural information of graphs, we use the Weisfeiler-Lehman(WL) isomorphism test \cite{Shervashidze2011WLkernel} as evaluation metrics to measure the similarity between graph structures.  The WL subtree kernel algorithm collects the information of neighbor nodes, and use hash to aggregate them to generate the label of the node in one iteration.
We devise a structural probe to explore whether these graph structural information are preserved in the graph embedding representation $\graphemb^k$.  The bottom-right figure (c) in Figure~\ref{fig:Overview}  shows the workflow of the structural probe.  The structural probe is to estimate whether the graph embedding has encoded the graph structural information e.g. the Weisfeiler-Lehman(WL) isomorphism test information. For the input graphs $\{\graph_1, \graph_2,\cdot,\graph_n\}$,  we obtain the graph representation $D=\{\graphemb_1^k, \graphemb_2^k, \cdots, \graphemb_n^k\}$ from the graph-based model \footnote{If the graph learning algorithm can learn the representation of the entire graph, we directly use it. Otherwise, we use the readout operation of nodes to represent the graph representation.}, and compute the pair-wise similarity of graph embeddings. Cosine similarity is used, and obtain the pairwise similarity of graph representations  $S^k_{cos}(\graph_m,\graph_n)$ for a pair of graphs $\graph_m$ and $\graph_n$:
\begin{equation}
\label{eqn:cosine}
    S^k_{cos}(\graph_m,\graph_n)=\frac{\graphemb^k_{\graph_m}\cdot\graphemb^k_{\graph_n}}{\Vert\graphemb^k_{\graph_m}\Vert\Vert\graphemb^k_{\graph_n}\Vert} \nonumber.
\end{equation}
And we also compute the  pair-wise similarity of the WL outputs for each pair of graphs. As the output of the WL algorithm is a set of each node labels, we use  Jaccard similarity to compute the structural-level graph similarity  $S^k_{str}(\graph_m,\graph_n)$ for a pair of graphs $\graph_m$ and $\graph_n$:
\begin{equation}
    S^k_{str}(\graph_m, \graph_n) = Jaccard(\wlabel(\graph_m), \wlabel(\graph_n)) \nonumber.
\end{equation}
$\wlabel(\graph_m)$ is the graph label output from the WL sub-tree kernel algorithm.
Based on the pair-wise similarity scores, we can have the pair-wise similarity matrix from the graph embedding level $\simmatrix^{GE}$ and the structural level $\simmatrix^{str}$.
Spearman correlation coefficient is used to estimate the structural scores $S^k_{s}$, by comparing the two similarity matrixes, computed as in Equation~\ref{eqn:structureprobe}.
\begin{equation}
\label{eqn:structureprobe}
    S^k_{s} = \frac{1}{n} \sum_{(\graph_m \in D}1-\frac{6\sum_{\graph n\in \dataset}\tau^2_{mn} }{n(n^2-1)}.
\end{equation}
$\tau_{mn}$ is the ordering distance between the ranked $R(\simmatrix^{GE}_m)$ and  the ranked $R(\simmatrix^{str}_m)$. $R(\cdot)$ is ranking operation on an input array.
A higher structure probe score $S^k_{s}$ indicates that the graph representations have the greater ability to capture the graph structural information. 

\section{Experiment} 
\label{sec:experiment}

\begin{table}
\centering
\caption{Benchmark datasets. The Classes indicate the node class number for the Cora and Flickr dataset, and graph class number for ENZYMES and MUTAG.}
\label{tab:datasets}
\resizebox{\linewidth}{!}{
\begin{tabular}{ccccccc}
\hline
                                                                           & Dataset   & Graphs & Classes & Nodes  & Edges  & Node Features \\ \hline
\begin{tabular}[c]{@{}c@{}}Citation\\ networks\end{tabular}                & Cora      & 1      & 7       & 2,708        & 5,429        & 1,433    \\ \hline
\multirow{3}{*}{\begin{tabular}[c]{@{}c@{}}Social\\ networks\end{tabular}} & Flickr    & 1      & 7       & 89,250       & 899,756      & 500      \\
                                                                           & Yelp      & 1      & -       & 69,716       & 1,561,406    & -        \\
                                                                           & MovieLens & 1      & -       & 10,352       & 100,836      & 404    \\ \hline
\multirow{2}{*}{\begin{tabular}[c]{@{}c@{}}Bio-\\ networks\end{tabular}}   & ENZYMES   & 600    & 6       & 33           & 124          & 21       \\
                                                                           & MUTAG     & 188    & 2       & 18           & 20           & 7        \\ \hline
\end{tabular}
}
\end{table}

\subsection{Representative Graph Learning Methods}
\label{sec:graphmodel}
In order to evaluate our probing methods on different categories, we use some representative graph learning methods for experimental evaluation and report their results in Section~\ref{sec:result} \footnote{Our probing method is flexibly used for any graph learning method. Due to space limitation, we chose different representative graph learning methods from diverse categories.}. In general, we select graph learning methods from 4 categories, including random walk based graph embedding methods (e.g. Node2Vec \cite{grover16:node2vec} and DeepWalk\cite{perozzi14:deepwalk}), basic graph neural networks (e.g.  GCN \cite{kipf17:gcn} and GAT \cite{velickovic17:gat}), self-supervised graph learning methods (e.g. GCL \cite{you:2020:GCL} and VGAE \cite{kipf2016VGAE}) and weighted graph learning methods (WGCN\cite{zhao2021wgcn}).
In addition, we add the control task, i.e. a naive  two-layer MLP method to avoid potential performance bias due to the learning performance of our probes \footnote{Our devised probes are based on MLP and we use MLP as the control experiment variable for comparison.}.
\begin{itemize}
\item \textbf{DeepWalk\cite{perozzi14:deepwalk}:} is one of random walk based graph embedding method. It adopts the local information obtained from truncated random walks to learn the latent representation of nodes via skip-Gram with hierarchical softmax.
\item \textbf{Node2Vec\cite{grover16:node2vec}:} is also a classic graph embedding method.
\item \textbf{Chebyshev\cite{defferrard16:cheby}:} generalizes convolutional neural networks (CNNs)  in the context of spectral graph theory and design fast localized convolutional filters on graphs for graph learning.
\item \textbf{GCN \cite{kipf17:gcn}:} performs semi-supervised learning on graph-structured data and introduces a simple and well-behaved layer-wise propagation rule for neural network models  via the localized first-order approximation of spectral graph convolutions. 
\item \textbf{GAT \cite{kipf17:gcn}:} incorporates masked self-attention layers on top of GCN-style methods.
\item \textbf{GraphSAGE \cite{hamilton17:sage}:} is an inductive framework for representation learning on large graphs which leverages node feature information to efficiently generate node embeddings for unseen data.
\item \textbf{VGAE\cite{kipf2016VGAE}:} is unsupervised learning framework based on the variational auto-encoder.
\item \textbf{GCL\cite{you:2020:GCL}:} is a graph contrastive learning framework for  unsupervised representation learning of graph data and devises four types of graph augmentations to incorporate various priors
\item \textbf{WGCN\cite{zhao2021wgcn}:} considers the directional structural information for different nodes and proposes a GCN model with weighted structural features.
\item \textbf{Control method (MLP):} we use a simple two layer MLP model with ReLu after the first layer as the control method.
\end{itemize}

 We perform probing evaluation on the representative graph learning methods for three classical downstream tasks of graph learning methods. In addition to learning with random initialization, we further study graph learning with meta-features initialized, and perform thorough analysis in Section~\ref{sec:result}.

\subsection{Downstream Tasks and Datasets}
 \label{sec:task}
 We conduct performance evaluation on the classic downstream tasks of graph learning methods, including node classification\cite{kipf17:gcn}, link prediction \cite{sen2008ngcf} and graph classification \cite{xu18:gin}.
\begin{itemize}
    \item \textbf{Node classification:} This task is one of the most popular and widely used applications of graph learning models. The graph learning methods learn the node representations and classify nodes into different groups.  
    \item \textbf{Link prediction:} It is to predicate whether there exist a link between two nodes. For example, the recommendation problem in recommender system scenarios can be formulated as one link prediction task and construct a user-item interaction graph to predict the probability of linking a user to a item. 
    \item \textbf{Graph classification:} Its goal is to classify a whole graph into different categories. The main applications of graph classification are protein classification and chemical compound classification. 
\end{itemize}

We adopt some benchmark datasets from different domains including citation networks, social networks, and Bio-chemical Networks. Table~\ref{tab:datasets} shows the statistics and proprietress of the used benchmarks.  
 \begin{itemize}
    \item \textbf{Citation Networks:} Cora\cite{mccallum2000cora} is a dataset containing scientific papers categorized into seven classes. It commonly be used for node classification (transductive). The number of node classes is 7, and the number of node features is 1,433.
    \item \textbf{Social Networks:} Flickr \cite{Zeng20:graphsaint} is a image hosting and video hosting platform. Flickr dataset is used for node classification (inductive).The number of node classes is 7 and the number of node features is 500. In addition, we adopt two social networks datasets Yelp \cite{sen2008ngcf} and MovieLens \cite{Harper15:movie} for Link prediction. Yelp includes 69,716 users and items as nodes, and 1,561,406 interaction records. MovieLens dataset includes 9,742 movies as nodes in the graph. It also includes over 100,836 ratings provided by around 610 users. The number of meta-features for items is 404.
   \item \textbf{Bio-Chemical Networks:} Two datsets are used for graph classification, respectively Enzyme \cite{Borgwardt05:proteins} and Mutag \cite{Debnath91:mutag}. Enzyme dataset contains the proteins information which contribute to catalyzing chemical reactions in the body. Mutag dataset is a widely used toxicity dataset which helps assess the potential risks of exposure to various chemicals and compounds. 
 \end{itemize}


\subsection{Experimental Setting}
All datasets expect Yelp dataset follow use the splitting rules as previous studies. For the Yelp dataset, the data splitting ratio for the training set, validation set and test set is set as 7:2:1. 

For Chebyshev, we use one layer structure. For GCN, GAT, GraphSAGE and MLP, we adopt the two layer structure. The length of walk for Node2Vec is set as 20.  The size of the context window for the skip-gram and the walks per node are both 10. In order to make the walk unbiased, we also set the p which controls how likely the walk is to go back to the previous node, it is set as 1.0. For WGCN method, we use the same set as \cite{zhao2021wgcn}, three layers structure, compute the neighbors number for each node and use the weighted neighbor features for each node. Also, we aggregate the weighted neighbor features for each node. Finally, we use ReLU activation function. In order to keep performance of it, we use the same encoder GCN as VGAE\cite{kipf2016VGAE}. For GCL, we use the same augmentation as the GCL, we randomly drop edges and swap some node in order to make different graphs. We use the cosine similarities to calculate the two augmenting graphs and calculate the contrastive loss. The activation function ELU is used for GAT and the activation function ReLU is used for the other models. The dropout ratio for all models is 0.5. The output dimension size for all models is 64. The learning rate is set as 0.001.

\section{Experimental Results} 
\label{sec:result}

\begin{table*}
\centering
\caption{Results of the graph learning models on Node Classification. Highlight the best performance with bold font and the worst ones with underlines.}
\label{tab:nodeperformance}
\resizebox{\linewidth}{!}{
\begin{tabular}{c|c|cccccccccc} 
\hline
Dataset-rand            & Metrics  & Chebyshev          & GCN                & GAT                & GraphSAGE & VGAE               & GCL                & WGCN               & Node2Vec           & DeepWalk           & MLP                 \\ 
\hline
\multirow{5}{*}{Cora}   & ACC      & 52.9 (6)           & 78.6 (3)           & \textbf{79.9 (1)}  & 77.3 (4)  & 79.7 (2)           & 45 (8)             & 76.13 (5)          & \uline{14.1 (10)}  & 21.24 (9)          & 52.4 (7)            \\
                        & F1       & 56.94 (6)          & 77.96 (3)          & \textbf{79.18 (1)} & 76.64 (4) & 78.84 (2)          & 45.7 (8)           & 74.32 (5)          & \uline{12.88 (10)} & 21.19 (9)          & 51.93 (7)           \\ 
\cline{2-12}
                        & EC       & 60.45 (7)          & 73.6 (2)           & \textbf{75.12 (1)} & 72.42 (4) & 72.83 (3)          & 62.38 (6)          & 67.17 (5)          & 57.5 (9)           & 58.13 (8)          & \uline{57.35 (10)}  \\
                        & BC       & 59.24 (6)          & 70.24 (3)          & \textbf{76.12 (1)} & 58.01 (7) & 71.24 (2)          & 61.13 (5)          & 64.13 (4)          & 54.1 (9)           & 54.12 (8)          & \uline{53.8 (10)}   \\ 
\cline{2-12}
                        & Distance & \uline{8.05 (10)}  & 11.24 (7)          & 12.98 (3)          & 11.93 (6) & \textbf{24.16 (1)} & 10.08 (9)          & 12.91 (4)          & 14.68 (2)          & 11.98 (5)          & 11.24 (7)           \\ 
\hline
\multirow{5}{*}{Flickr} & ACC      & 77.5 (5)           & 81.14 (4)          & 72.05 (6)          & 57.05 (7) & 91.41 (2)          & \textbf{92.8 (1)}  & 85.12 (3)          & 50.49 (8)          & 50.12 (9)          & \uline{47.94 (10)}  \\
                        & F1       & 77.5 (5)           & 81.14 (4)          & 72.05 (6)          & 57.05 (7) & 91.41 (2)          & \textbf{92.8 (1)}  & 85.12 (3)          & 50.49 (8)          & 50.12 (9)          & \uline{47.94 (10)}  \\ 
\cline{2-12}
                        & EC       & 49.7 (7)           & 50 (6)             & 50.05 (4)          & 50.05 (4) & 71.28 (2)          & \textbf{76.12 (1)} & 69.12 (3)          & 32.18 (8)          & 31.24 (9)          & \uline{23.13 (10)}  \\
                        & BC       & 68.17 (4)          & 65.72 (7)          & 66.12 (6)          & 67.12 (5) & 70.13 (2)          & \textbf{75.13 (1)} & 70.1 (3)           & 30.13 (8)          & \uline{28.17 (10)} & 29.38 (9)           \\ 
\cline{2-12}
                        & Distance & 13.45 (6)          & 13.82 (2)          & 13.82 (2)          & 12.54 (7) & 13.62 (5)          & 13.82 (2)          & \textbf{19.52 (1)} & 11.98 (8)          & 10.44 (9)          & \uline{9.68 (10)}   \\ 
\hline
Dataset-meta            & \multicolumn{11}{c}{Initialization with meta-features}                                                                                                                                                             \\ 
\hline
\multirow{5}{*}{Cora}   & ACC      & 62.3 (7)           & 79.4 (2)           & \textbf{79.9 (1)}  & 77.3 (5)  & 78.5 (3)           & 55.14 (9)          & 78.12 (4)          & 55.18 (8)          & 68.91 (6)          & \uline{53.4 (10)}   \\
                        & F1       & 63.9 (6)           & 78.57 (2)          & \textbf{79.02 (1)} & 76.52 (5) & 77.83 (4)          & 55.18 (9)          & 78.01 (3)          & 58.13 (8)          & 61.13 (7)          & \uline{52.8 (10)}   \\ 
\cline{2-12}
                        & EC       & 65.79 (6)          & \textbf{72.37 (1)} & 72.14 (2)          & 66.18 (4) & 70.24 (3)          & 41.23 (8)          & 66.01 (5)          & 52.37 (7)          & 32.44 (9)          & \uline{30.12 (10)}  \\
                        & BC       & 63.71 (6)          & \textbf{70.12 (1)} & 69.27 (2)          & 66.04 (4) & 66.13 (3)          & 51.28 (8)          & 64.78 (5)          & 57.89 (7)          & 41.28 (9)          & \uline{39.38 (10)}  \\ 
\cline{2-12}
                        & Distance & \uline{8.03 (10)}  & 11.18 (8)          & 13.2 (4)           & 12.03 (5) & 11.47 (6)          & 11.04 (9)          & \textbf{30.91 (1)} & 14.59 (2)          & 13.82 (3)          & 11.19 (7)           \\ 
\hline
\multirow{5}{*}{Flickr} & ACC      & 77.9 (5)           & 82.12 (4)          & 74.13 (6)          & 59.38 (7) & 92.42 (2)          & \textbf{93.12 (1)} & 86.13 (3)          & 52.48 (8)          & 52.34 (9)          & \uline{50.13 (10)}  \\
                        & F1       & 77.9 (5)           & 82.12 (4)          & 74.13 (6)          & 59.38 (7) & 92.42 (2)          & \textbf{93.12 (1)} & 86.13 (3)          & 52.48 (8)          & 52.34 (9)          & \uline{50.13 (10)}  \\ 
\cline{2-12}
                        & EC       & 48.37 (7)          & 81.38 (5)          & 68.73 (6)          & 86.81 (3) & 90.38 (2)          & \textbf{91.24 (1)} & 85.12 (4)          & 48.37 (7)          & 46.37 (9)          & \uline{42.41 (10)}  \\
                        & BC       & 38.14 (6)          & 39.14 (4)          & 37.89 (7)          & 38.84 (5) & 67.4 (2)           & \textbf{68.38 (1)} & 59.24 (3)          & 32.32 (8)          & 29.38 (9)          & \uline{29.04 (10)}  \\ 
\cline{2-12}
                        & Distance & \textbf{24.25 (1)} & \textbf{24.25 (1)} & 21.48 (5)          & 19.09 (7) & 23.01 (4)          & 18.33 (8)          & 23.62 (3)          & 18.02 (9)          & 19.1 (6)           & \uline{16.04 (10)}  \\
\hline
\end{tabular}
}
\end{table*}

\begin{table*}
\centering
\caption{Results of the graph learning models on Link Prediction. Highlight the best performance with bold font and the worst ones with underlines.}
\label{tab:linkperformance}
\resizebox{\linewidth}{!}{
\begin{tabular}{c|c|cccccccccc} 
\hline
Dataset-rand                 & Metrics  & Chebyshev & GCN                & GAT                & GraphSAGE          & VGAE               & GCL                & WGCN               & Node2Vec  & DeepWalk  & MLP                 \\ 
\hline
\multirow{5}{*}{Yelp}        & AUC      & 61.24 (7) & \textbf{78.14 (1)} & 63.12 (6)          & 63.71 (5)          & 73.14 (3)          & 67.12 (4)          & 76.12 (2)          & 55.12 (8) & 52.37 (9) & \uline{42.32 (10)}  \\
                             & F1       & 52.24 (8) & \textbf{71.50 (1)} & 66.15 (5)          & 65.23 (6)          & 71.09 (2)          & 70.01 (3)          & 68.57 (4)          & 53.17 (7) & 51.95 (9) & \uline{51.05 (10)}  \\ 
\cline{2-12}
                             & Distance & 31.90 (3) & \textbf{32.02 (1)} & 24.25 (5)          & 22.05 (8)          & 24.25 (5)          & 28.32 (4)          & 32.01 (2)          & 24.18 (7) & 16.33 (9) & \uline{14.04 (10)}  \\ 
\cline{2-12}
                             & EC       & 46.28 (9) & 71.21 (2)          & 60.71 (5)          & 64.28 (4)          & \textbf{72.16 (1)} & 56.73 (6)          & 70.12 (3)          & 49.82 (7) & 48.12 (8) & \uline{45.39 (10)}  \\
                             & BC       & 45.13 (7) & \textbf{72.12 (1)} & 56.78 (6)          & 63.47 (4)          & 71.97 (2)          & 58.12 (5)          & 71.12 (3)          & 39.02 (9) & 40.18 (8) & \uline{36.85 (10)}  \\ 
\hline
\multirow{5}{*}{MovieLens~~} & AUC      & 68.21 (5) & 73.75 (2)          & \uline{47.91 (10)} & 71.21 (3)          & \textbf{74.29 (1)} & 68.14 (6)          & 70.12 (4)          & 56.43 (7) & 51.24 (8) & 50.12 (9)           \\
                             & F1       & 50.75 (9) & 67.94 (3)          & 61.25 (4)          & 58.64 (5)          & 70.48 (2)          & 58.60 (6)          & \textbf{70.87 (1)} & 51.84 (7) & 51.06 (8) & \uline{48.82 (10)}  \\ 
\cline{2-12}
                             & Distance & 19.50 (8) & 19.52 (7)          & 29.78 (2)          & 24.25 (4)          & 24.17 (6)          & 28.77 (3)          & \textbf{31.82 (1)} & 24.20 (5) & 15.75 (9) & \uline{12.31 (10)}  \\ 
\cline{2-12}
                             & EC       & 50.52 (7) & 70.85 (2)          & 61.23 (5)          & 63.41 (4)          & \textbf{72.16 (1)} & 59.13 (6)          & 70.12 (3)          & 49.82 (9) & 50.12 (8) & \uline{41.26 (10)}  \\
                             & BC       & 40.97 (8) & \textbf{72.30 (1)} & 57.55 (6)          & 63.77 (4)          & 71.97 (2)          & 62.12 (5)          & 69.13 (3)          & 39.02 (9) & 42.12 (7) & \uline{37.24 (10)}  \\ 
\hline
Dataset-meta                 & \multicolumn{11}{c}{Initialization with meta-features}                                                                                                                                           \\ 
\hline
\multirow{5}{*}{Yelp}        & AUC      & 63.85 (7) & 81.14 (2)          & 78.03 (4)          & 70.37 (5)          & 78.24 (3)          & 70.13 (6)          & \textbf{82.34 (1)} & 59.12 (8) & 55.37 (9) & \uline{45.32 (10)}  \\
                             & F1       & 53.44 (9) & \textbf{75.35 (1)} & 67.45 (5)          & 66.83 (6)          & 73.19 (4)          & 74.60 (2)          & 73.89 (3)          & 57.12 (7) & 54.82 (8) & \uline{52.44 (10)}  \\ 
\cline{2-12}
                             & Distance & 28.96 (4) & 33.20 (2)          & 24.18 (8)          & 24.47 (7)          & 24.92 (6)          & 25.70 (5)          & \textbf{49.06 (1)} & 29.87 (3) & 19.47 (9) & \uline{13.82 (10)}  \\ 
\cline{2-12}
                             & EC       & 47.21 (8) & 72.43 (2)          & 61.24 (5)          & 65.12 (4)          & \textbf{73.13 (1)} & 57.14 (6)          & 71.21 (3)          & 50.12 (7) & 47.12 (9) & \uline{42.31 (10)}  \\
                             & BC       & 46.14 (7) & 73.13 (3)          & 57.38 (6)          & 64.18 (4)          & \textbf{73.59 (1)} & 59.12 (5)          & 73.24 (2)          & 41.31 (9) & 45.89 (8) & \uline{38.13 (10)}  \\ 
\hline
\multirow{5}{*}{MovieLens~}  & AUC      & 70.57 (4) & 73.56 (3)          & \uline{48.37 (10)} & 74.92 (2)          & \textbf{74.96 (1)} & 52.38 (8)          & 70.18 (5)          & 56.66 (7) & 57.17 (6) & 50.12 (9)           \\
                             & F1       & 62.45 (5) & 66.16 (3)          & 45.63 (8)          & \textbf{71.86 (1)} & 67.97 (2)          & 44.22 (9)          & 64.51 (4)          & 51.65 (6) & 50.13 (7) & \uline{43.59 (10)}  \\ 
\cline{2-12}
                             & Distance & 27.15 (5) & 29.77 (2)          & 22.43 (7)          & 28.02 (4)          & 22.84 (6)          & 28.68 (3)          & \textbf{32.28 (1)} & 20.51 (8) & 19.52 (9) & \uline{16.33 (10)}  \\ 
\cline{2-12}
                             & EC       & 60.73 (5) & \textbf{72.63 (1)} & 71.88 (2)          & 60.36 (6)          & 71.61 (3)          & \uline{31.35 (10)} & 65.35 (4)          & 55.14 (8) & 60.12 (7) & 42.31 (9)           \\
                             & BC       & 53.29 (6) & 71.10 (2)          & \textbf{75.79 (1)} & 50.47 (8)          & 71.04 (3)          & 43.13 (9)          & 67.12 (4)          & 53.17 (7) & 58.38 (5) & \uline{40.13 (10)}  \\
\hline
\end{tabular}

}
\end{table*}

\begin{table*}
\centering
\caption{Results of the graph learning models on Graph Classification. Highlight the best performance with bold font and the worst ones with underlines.}
\label{tab:graphperformance}
\resizebox{\linewidth}{!}{
\begin{tabular}{c|c|cccccccccc} 
\hline
Dataset-rand             & Metrics   & Chebyshev          & GCN                & GAT                & GraphSAGE & VGAE               & GCL                & WGCN               & Node2Vec  & DeepWalk  & MLP                 \\ 
\hline
\multirow{3}{*}{MUTAG}   & ACC       & 82.05 (5)          & \textbf{92.31 (1)} & 87.18 (2)          & 84.62 (4) & 74.36 (6)          & 74.36 (6)          & 87.13 (3)          & 72.71 (9) & 73.13 (8) & \uline{71.49 (10)}  \\
                         & F1        & 82.05 (5)          & \textbf{92.31 (1)} & 87.18 (2)          & 84.62 (4) & 74.36 (6)          & 74.36 (6)          & 87.13 (3)          & 72.71 (9) & 73.13 (8) & \uline{71.49 (10)}  \\ 
\cline{2-12}
                         & Structure & 11.96 (4)          & \textbf{15.69 (1)} & 13.39 (2)          & 8.29 (5)  & 3.89 (7)           & 3.79 (8)           & 12.23 (3)          & 5.12 (6)  & 3.13 (9)  & \uline{2.62 (10)}   \\ 
\hline
\multirow{3}{*}{ENZYMES} & ACC       & 45 (4)             & 44.17 (5)          & 45.83 (3)          & 54.17 (2) & 43.33 (6)          & 43.33 (6)          & \textbf{59.38 (1)} & 42.5 (8)  & 41.5 (9)  & \uline{37.5 (10)}   \\
                         & F1        & \uline{15 (10)}    & 22.5 (3)           & 20.83 (4)          & 18.33 (5) & \textbf{26.67 (1)} & 16.67 (8)          & 25.81 (2)          & 18.27 (6) & 17.93 (7) & 16.67 (8)           \\ 
\cline{2-12}
                         & Structure & 3.44 (7)           & 11.82 (3)          & 8.37 (4)           & 12.83 (2) & 7.26 (6)           & 8.36 (5)           & \textbf{15.48 (1)} & 2.19 (8)  & 2.04 (9)  & \uline{1.84 (10)}   \\ 
\hline
Dataset-meta             & \multicolumn{11}{c}{Initialization with meta-features}                                                                                                                                            \\ 
\hline
\multirow{3}{*}{MUTAG}   & ACC       & \textbf{94.87 (1)} & 82.05 (6)          & 84.62 (4)          & 92.31 (2) & 79.49 (7)          & 84.62 (4)          & 90.24 (3)          & 77.31 (8) & 75.13 (9) & \uline{74.62 (10)}  \\
                         & F1        & \textbf{94.87 (1)} & 82.05 (6)          & 84.62 (4)          & 92.31 (2) & 79.49 (7)          & 84.62 (4)          & 90.24 (3)          & 77.31 (8) & 75.13 (9) & \uline{74.62 (10)}  \\ 
\cline{2-12}
                         & Structure & \textbf{18.19 (1)} & 4.78 (7)           & 6.02 (5)           & 16.48 (3) & 5.98 (6)           & 15.8 (4)           & 17.29 (2)          & 2.38 (9)  & 2.41 (8)  & \uline{1.73 (10)}   \\ 
\hline
\multirow{3}{*}{ENZYMES} & ACC       & 43.33 (5)          & 47.5 (3)           & \uline{39.17 (10)} & 57.5 (2)  & 43.33 (5)          & 46.67 (4)          & \textbf{60.35 (1)} & 41.32 (8) & 40.13 (9) & 43.33 (5)           \\
                         & F1        & 22.5 (3)           & 21.67 (4)          & \uline{15 (10)}    & 20 (5)    & 19.17 (6)          & \textbf{25.83 (1)} & 25.23 (2)          & 17.99 (8) & 18.12 (7) & 17.5 (9)            \\ 
\cline{2-12}
                         & Structure & 7.49 (3)           & 5.12 (4)           & 3.48 (6)           & 8.89 (2)  & 3.24 (7)           & 4.56 (5)           & \textbf{10.23 (1)} & 1.31 (9)  & 1.78 (8)  & \uline{0.39 (10)}   \\
\hline
\end{tabular}
}
\end{table*}

\begin{figure*}[htbp]
\centering
\includegraphics[width=0.98\linewidth]{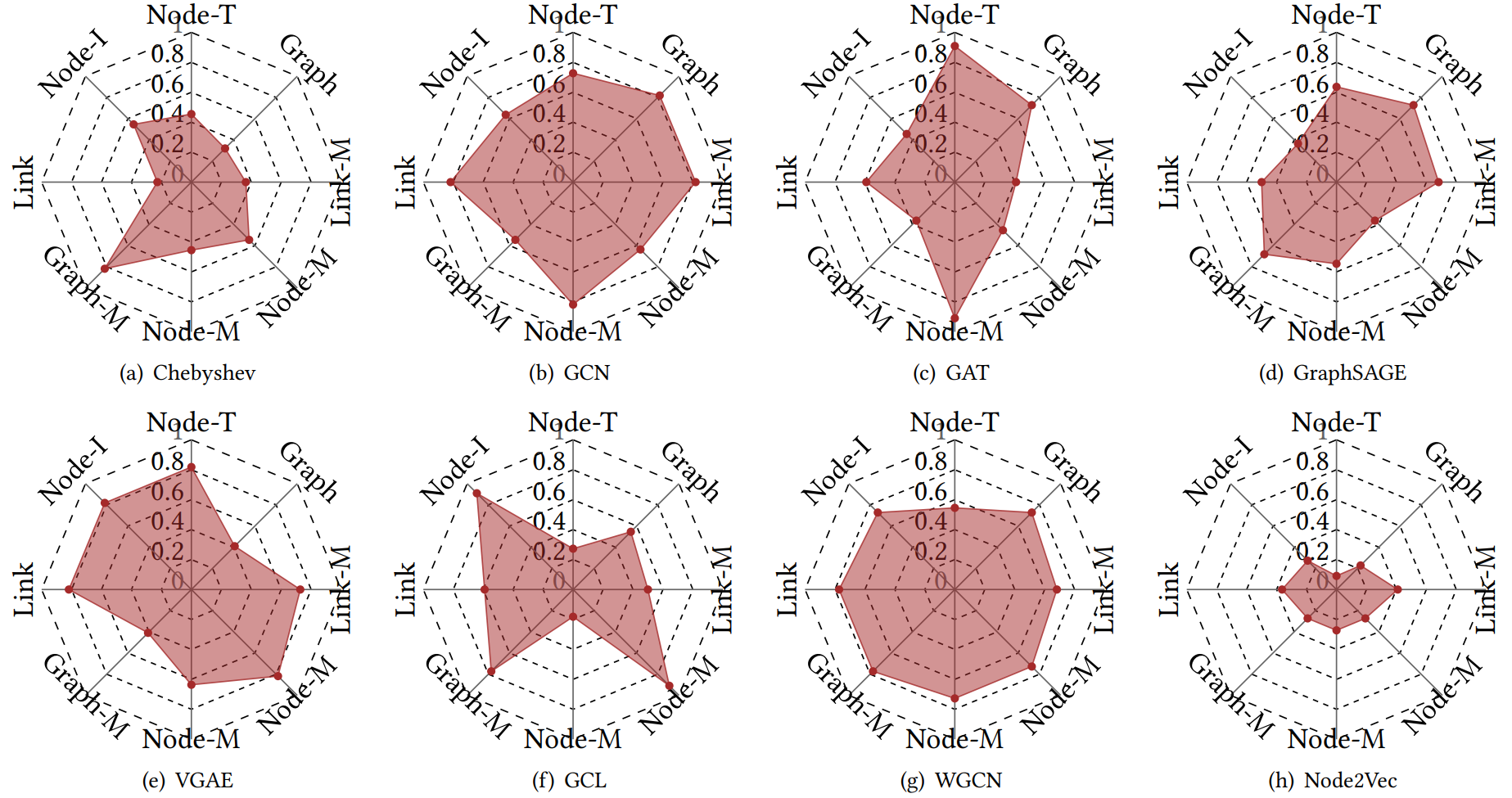}
\caption{Radar chart comparison of graph embedding models for different information embedding capabilitie}
\label{fig:radar}
\Description{Radar}
\end{figure*}

\begin{figure*}[htbp]
\centering
\subfigure{
\includegraphics[width=0.32\linewidth]{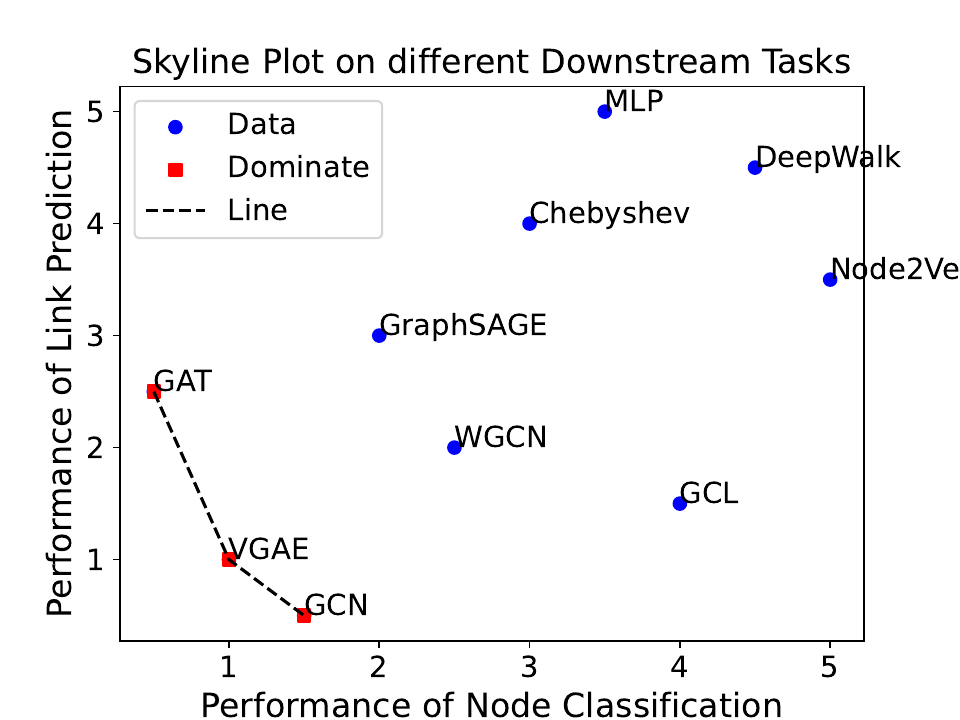}
\includegraphics[width=0.32\linewidth]{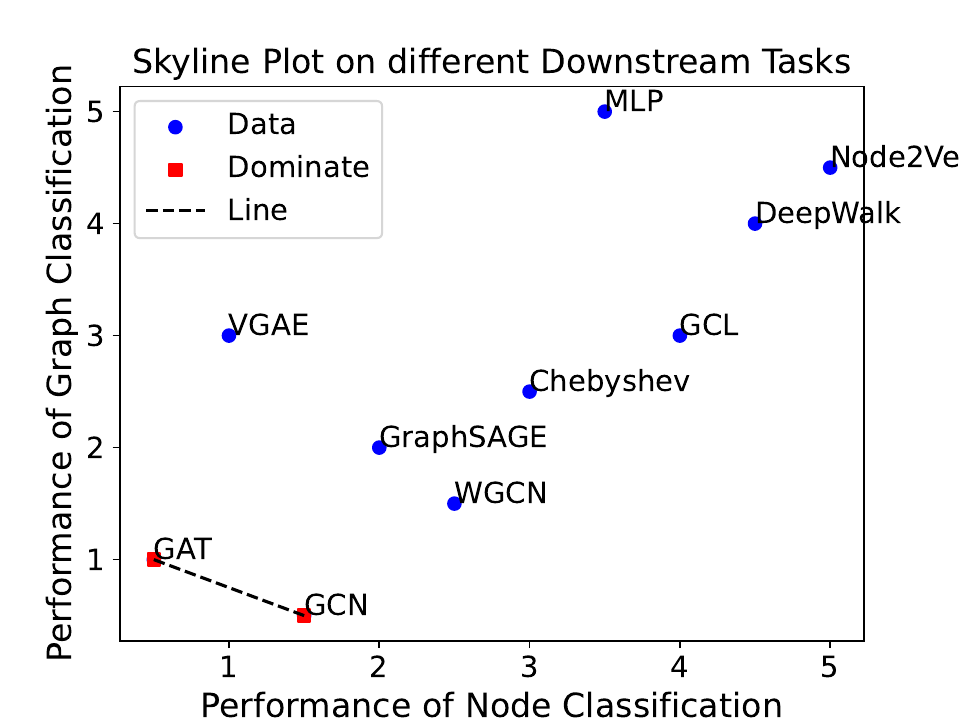}
\includegraphics[width=0.32\linewidth]{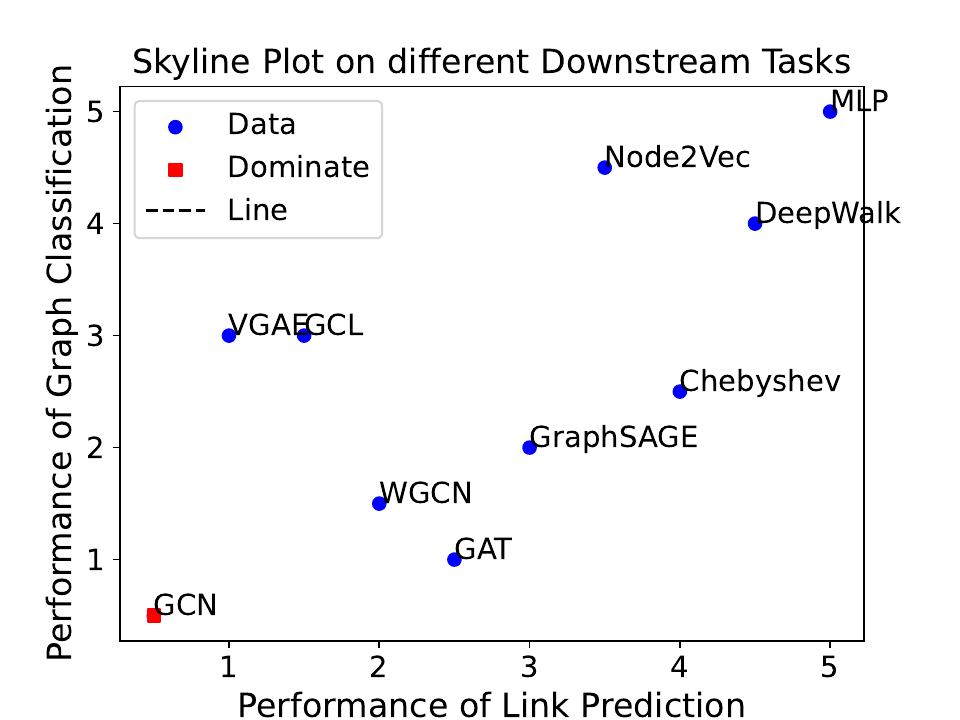}
}

\caption{Skyline Plot on different Downstream Tasks based on random initialization }
\label{fig:skyline}
\Description{Skyline}
\end{figure*}

\subsection{Performance of Graph Learning Methods on Node Classification}
In order to validate the knowledge probing performance of our methods with respect to the node classification task, we compare the probing scores of the representative graph learning methods with reference to commonly used metrics in node classification including Accuracy (ACC) and F1 scores. Table~\ref{tab:nodeperformance} shows the performance comparison for node classification. 
In additional to the absolute performance numbers, we add the overall rank numbers among the compared methods with brackets as the relative performance. For example, the highlighted number 79.9(1) indicates GAT obtains 0.799 accuracy on Cora dataset, and rank first among all compared methods. In all tables, we use bold font to highlight the best performance and underlines to indicate the worst performance \footnote{ Statistical significance tests have been conducted. All reported results are significant}.

For node classification, we report transductive performance from Cora and inductive performance from Flickr. We can see that the best method on Cora (transductive) is GAT, and the best method is GCL on Flick (inductive) with respect to both ACC and F1 scores.
Our centrality probes (EC and BC) have consistent evaluation results with the commonly used golden metrics, namely GAT ranks first. 
For the worst cases, our centrality prober has the same results (DeepWalk or MLP) with ACC and F1 on Flick dataset. On Cora, we find an interesting phenomenon that the simple MLP methods is better than GCL, Node2Vec and DeepWalk with respect to Acc and F1 while our centrality probing method demonstrate the worst methods are respectively MLP and Node2Vec. It might raise the question whether the final statistical metrics like accuracy and F1 actually reflect the capability of the graph representation learning 

In comparison with the centrality probes, our distance probe are not very consistent with our centrality probes and the traditional metrics although they can find the worse cases. It is because the the distance probe is a path-wise probe while the node classification might emphasize to encode more topological information into graph representation learning. The path-wise probe method e.g. the distance probe might be inappropriate for knowledge probing in the node classification task as only encoding the path-wise information in graph representation learning cannot works well, e.g. random walk based methods like DeepWalk and Node2Vec.

When initializing with meta-features, we have the similar results on Flick dataset, and the best method is GCL with respect to both of our centrality probes and the classical metrics. On Cora, the top-2 method with respect to ACC and F1 are GAT and GCN while top-2 methods with respect to our centrality probes are GAT and CCN. On Cora and Flick datasets, the worst cases are all MLP with respect to both of our centrality probes and the classical metrics. In general, it has slight perturbation in the relative performance when initialized with meta-features.

In order to further discuss the relations between the node-wise properties and the graph representation learning, we calculate the homophily \cite{zhu2020beyond} of different graphs from different datasets. The homophily ratio of is about 0.8 for Cora nad 0.32 for Flickr.  It might explain why different graph methods show different performance, due to the inherent graph properties. GNN methods like GAT can have better performance on highly homophilous graphs. Our probe also have consistent results with the homophily analysis.
\begin{itemize}
    \item  \textbf{Our centrality probes is effective for knowledge probing of graph representation learning in the node classification task.} They has consistent results with the traditional evaluation metrics accuracy and F1, validating their effectiveness of our centrality probes. 
    \item The path-wise probe method e.g. the distance probe might be inappropriate for knowledge probing in the node classification task. 
    \item \textbf{GAT is superior to other representative methods.} The top-2 methods are GAT and GCN. Initializing with meta-measures results into small perturbation among the top percentile in performance ranking.
    
\end{itemize}

\subsection{Performance of Graph Learning Methods on Link Prediction}
To evaluate the knowledge probing performance of our methods with respect to the link prediction task, we compare the probing scores of the representative graph learning methods with reference to commonly used metrics in the link prediction task on Yelp and Movielens dataset, including AUC and F1 scores. We evaluate both the centrality probes and the distance probes, and the performance comparison results on Yelp and MovieLens for link prediction are reported in Table~\ref{tab:linkperformance}. On Yelp dataset, the best methods is GCN and the worst method is DeepWalk except MLP, with respect to both AUC and F1 scores. The distance probe has consistent results with both AUC and F1 scores. Although the centrality probes cannot be totally the same with the traditional scores, we can see that they ( especially the centrality probe with betweeness) have similar results by ranking GCN, VGAE at top-3 positions in the case that the two traditional metrics have similar results rather than the consistent results. On Moivelens dataset, the traditional metrics have shown considerable different results in the performance ranking, in which the best method is VGAE with AUC and that is WGCN with F1. This phenomenon to some extent indicates that different evaluation metrics might have some different relative results due to different measure mechanisms. Our distance probe is consistent with F1 for the best method, and our centrality probes are more similar with AUC scores.

When initializing with meta-features, the performance of link prediction on Yelp and Movielens have some different results. The best method on Yelp is WGCN (AUC) and GCN (F1), in contrast with random initialization (GCN). On Moivelens dataset, the best ones are VGAE (AUC) and GraphSAGE (F1) while the random initialization results are VGAE (AUC) and WGCN (F1). Our distance probe also capture the differences due to different initialization setting and is still consistent with the traditional metrics, ranking WGCN and GCN at the top-2 positions.

\begin{itemize}
    \item  \textbf{The distance probe is effective for knowledge probing of graph representation learning in the link prediction task.} They have consistent results with the traditional evaluation metrics AUC and F1 for different initialization setting. The reason might be that the path-wise probe is devised to probe the path-wise information within graph representation learning which acts the core roles in link prediction.
    \item In general, no single method can be totally superior to other methods with respect to all evaluation metrics on the two datasets. GCN, VGAE and WGCN can be ranked at the top positions.
\end{itemize}

\subsection{Performance of Graph Learning Methods on Graph Classification} 
We compare the structure probing scores of the representative graph learning methods with reference to commonly used metrics in the graph classification task on MUTAG and ENZYMES dataset, including accuracy (ACC) and F1 scores. Table~\ref{tab:linkperformance} demonstrates the performance results on graph classification.
On MUTAG dataset, the best method is GCN with with both ACC and F1 and our structural probing result is consistent with the traditional golden metrics. On ENZYMES dataset, our structural probing results are consistent with ACC scores (WGCN is best).  For the cases with meta-features, our structural probing results are consistent with ACC and F1 scores on MUTAG dataset and with ACC scores (Chebyshev is best) on ENZYMES dataset (WGCN is best). 

\begin{itemize}
    \item  \textbf{Our structure probe is effective for knowledge probing of graph representation learning in the graph classification task.} They have consistent results with the traditional evaluation metrics ACC and F1 for different initialization setting. 
    \item \textbf{In general, no single method can dominate other methods on the two datasets.} WGCN has relatively robust peformance, ranking at top-3 positions on the two datasets. In some cases, some "out-of-date" methods e.g. Chebyshev can have better performance.
\end{itemize}


\subsection{Visualization Analysis}
In order to further compare the overall performance of different methods for different information embedding capabilities, we compute the ranking 
 of the 9 representative methods and the MLP baseline for each probe and use radar charts to visualize their capacities in Figure~\ref{fig:radar} (I indicates the inductive and T indicates transductive, and M the Meta) . \textbf{GCN and WGCN have better performance in most probing aspects in comparison with other methods.} Although having the same capacities of aggregating the neighbor information, VGAE has the sub-optimal performance. Furthermore, they all hardly rely on meta informations. GCL has better performance on Node Classification (Inductive), GAT has better performance on Node Classification(transductive). Chebyshev might be better used with meta information. Node2Vec has the worst performance on all aspects. 
 
 We also draw the skyline plot for finding the methods on different downstream tasks which can not be dominated by other methods in Figure ~\ref{fig:skyline}. We can investigate the joint-abilities of graph learning methods on different downstream tasks. In Link Prediction-Node Classification tasks, GAT, VGAE and GCN has better performance that can not be dominated. In Graph Classification-Node Classification, GAT and GCN performs well. GAT has better graph classification capabilities and GCN has better node classification abilities. In Graph Classification-Link Predictions, only GCN cannot be dominated by others. From the skyline results, we can see that GCN and GAT has better joint-abillites for downstream tasks. 
 
\subsection{Effects of Parameters}
The only hyper parameters used in our probes is the path parameter. It controls the shortest path that our probe can detect in the graph. We calculate the Correlations with the F1 scores in different path, it shows that most of our best path are between 3 and 4. We use the max score of the path for each datasets

\section{Conclusion} \label{conlusion}
In this paper, we proposed a graph probing benchmark for the representative graph learning methods. Diverse probes at three different levels (node-wise, path-wise and structure-wise) are devised to investigate and interpret weather the graph properties from different levels are encoded into representation learning of the seven representative graph neural networks based methods. We conduct system evaluation and thorough analysis to investigate what kind of information have been encoded and which methods have competitive performance with different targeted downstream tasks. The experimental evaluation validate the effectiveness of GraphProbe. Furthermore, We conclude some remarking findings: GAT is superior in the node classification; GCN and WGCN are relatively versatile methods achieving better results with respect to different tasks. 
The benchmark codes and resources will be public after acceptance.

\label{sec:A-GNN}

\bibliography{anthology,custom}
\bibliographystyle{acl_natbib}

\end{document}